\documentclass[10pt, a4paper]{article}
\usepackage{lrec}
\usepackage{xcolor}
\usepackage{amssymb}
\usepackage{dsfont}

\usepackage{graphicx}
\usepackage{tabularx}
\usepackage{soul}
\usepackage{amsmath}
\usepackage{enumitem}
\usepackage{listings}
\usepackage{xcolor}

\usepackage{url}

\definecolor{codegreen}{rgb}{0,0.6,0}
\definecolor{codegray}{rgb}{0.5,0.5,0.5}
\definecolor{codepurple}{rgb}{0.58,0,0.82}
\definecolor{backcolour}{rgb}{0.95,0.95,0.92}
 
\lstdefinestyle{mystyle}{
    backgroundcolor=\color{backcolour},   
    commentstyle=\color{codegreen},
    keywordstyle=\color{magenta},
    numberstyle=\tiny\color{codegray},
    stringstyle=\color{codepurple},
    basicstyle=\ttfamily\footnotesize,
    breakatwhitespace=false,         
    breaklines=true,                 
    captionpos=b,                    
    keepspaces=true,                 
    numbers=left,                    
    numbersep=5pt,                  
    showspaces=false,                
    showstringspaces=false,
    showtabs=false,                  
    tabsize=2
}
 
\usepackage{epstopdf}
\usepackage[utf8]{inputenc}

\usepackage{hyperref}
\usepackage{xstring}
\newcommand*\samethanks[1][\value{footnote}]{\footnotemark[#1]}

\definecolor{darkgreen}{rgb}{0.0, 0.2, 0.13}
\definecolor{britishracinggreen}{rgb}{60,179,113}

\usepackage{relsize}

\title{Seshat: A tool for managing and verifying annotation campaigns of audio data}

\name{Hadrien Titeux$^{\star 1}$ \thanks{$\star$ Equal contribution. 
This work was conducted while E. Dupoux was a part-time Research Scientist at Facebook AI Research.
Code for Seshat is available on Github at \url{https://github.com/bootphon/seshat}}, Rachid Riad$^{\star 1,2}$\samethanks[1], Xuan-Nga Cao$^{1}$,  Nicolas  Hamilakis$^{1}$, Kris Madden$^{1}$, \\
 {\bf \large Alejandrina Cristia$^{1}$,  Anne-Catherine Bachoud-Lévi$^{2}$, Emmanuel Dupoux$^{1}$}}

\address{
  $1$ LSCP/ENS/CNRS/EHESS/INRIA/PSL Research University, Paris, France \\
  $2$ NPI/ENS/INSERM/UPEC/PSL Research University, Cr\'eteil, France  \\
         \{hadrien.titeux, rachid.riad\}@ens.fr,
         \\
         \{ngafrance, nick.hamilakis562, thekrismadden, alecristia,  bachoud,  emmanuel.dupoux\}@gmail.com}

\abstract{
We introduce Seshat, a new, simple and open-source software to efficiently manage annotations of speech corpora. The Seshat software allows users to easily customise and manage annotations of large audio corpora while ensuring compliance with the formatting and naming conventions of the annotated output files. In addition, it includes procedures for checking  the content of annotations following specific rules that can be implemented in personalised parsers. Finally, we propose a double-annotation mode, for which Seshat computes automatically an associated inter-annotator agreement with the $\gamma$ measure taking into account the categorisation and segmentation discrepancies. 
\\ \newline  }

\begin{document}

\maketitleabstract

\section{Introduction}
 Large corpora of speech, obtained in the laboratory and in naturalistic conditions, become easier to collect. This new trend broadens the scope of scientific questions on speech and language that can be answered. However, this poses an important challenge for the construction of reliable and usable annotations. Managing annotators and ensuring the quality of their annotations are highly demanding tasks for research endeavours and industrial projects \cite{zue1990speech}. When organised manually, the manager of annotation campaigns usually faces three major problems: the \textit{mishandling of files} (e.g., character-encoding problems, incorrect naming of files), the \textit{non-conformity of the annotations} \cite{moreno2000speechdat}, and the \textit{inconsistency of the annotations} \cite{gut2004measuring}. 

In this paper, we introduce \textit{Seshat}, a system for the automated management of annotation campaigns for audio/speech data which addresses these challenges. It is built on two components that communicate via a Restful API: a back-end (server) written in Flask and a front-end (client) in Angular Typescript. Seshat is easy to install for non-developers and easy to use for researchers and annotators while having some extension capabilities for developers.

In Section \ref{main:rel}, we describe the related work on annotations tools, which do not provide solutions to all the aforementioned challenges during corpus creation.  In Section \ref{main:overview}, we make an overview of the different functionalities of the software. Then, we explain, in Section \ref{main:dev}, the architecture of the software, and also the several UX/UI design and engineering choices that have been made to facilitate the usage of the platform. We describe how to use of Seshat in Section \ref{main:using} and Section \ref{main:use_cases} presents two specific use-cases. Finally, we conclude and describe future plans for Seshat in Section \ref{main:conclusion}.

\section{Related Work}
\label{main:rel}
\textbf{Self-hosted annotation systems.} There are many standalone solutions for the transcription of speech data that are already used by researchers: Transcriber \cite{barras2001transcriber}, Wavesurfer \cite{sjolander2000wavesurfer}, Praat \cite{boersma2002praat}, ELAN \cite{macwhinney2014childes}, XTrans \cite{glenn2009xtrans}. These systems allow the playback of sound data and the construction of different layers of annotations with various specifications, with some advanced capabilities (such as annotations with hierarchical or no relationship between layers, number of audio channels, video support).
Yet, these solutions lack a management system: each researcher must track the files assigned to annotators and build a pipeline to parse (and eventually check) the output annotation files. Moreover, checking can only be done once the annotations have been submitted to the researchers. This task becomes quickly untraceable as the number of files and annotators grow.  
In addition, most of these transcription systems do not provide a way to evaluate consistency (intra- and inter-annotator agreement) that would be appropriate for speech data \cite{mathet2015unified}. 

\textbf{Web-based annotations systems.} 
There are several web-based annotation systems for the annotation of audio data. Among them we find light-weight systems, like the VIA software \cite{dutta2019via} or Praat on the web \cite{dominguez2016praat} that allow to build simple layers of annotations. However, they do not provide a proper management system for a pool of annotators nor do they integrate annotation checking. 

On the other side of the spectrum, there are more sophisticated systems with various capabilities. Camomille \cite{poignant2016camomile} and the EMU-SDMS system (that can also be used offline) \cite{winkelmann2017emu} allow to work with speech data and to distribute the tasks to several annotators. But these systems require expertise in web hosting and technologies to deploy and modify them.

Finally, WebAnno \cite{yimam2013webanno} and GATE Teamware \cite{bontcheva2013gate} are the tools that most closely match our main contributions regarding quality control (conformity and consistency checking), annotators' management and flexibility. WebAnno includes consistency checking with the integration of different metrics \cite{meyer2014dkpro}. However, these tools have only been built for text data. The format and all the custom layers have been designed for Natural Language Processing tasks. Porting WebAnno to support speech data seemed a major engineering challenge. That is why it appeared necessary to develop a new and user-friendly tool addressed to the speech community.



\section{Overview of Seshat}
\label{main:overview}
\begin{figure}[!ht]
  \includegraphics[width=0.48\textwidth]{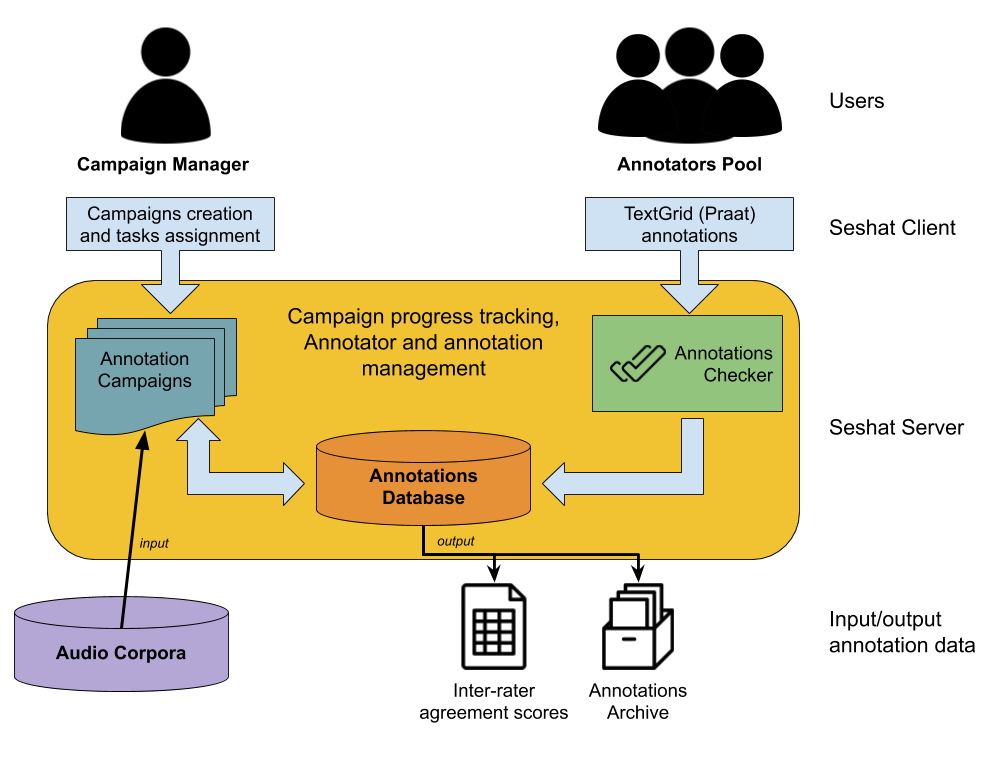}
  \caption{Seshat architecture: two different front-end, for annotators and campaign manager, a back-end with persistent data storage of the annotations and the inter-rater agreements.}
  \label{fig:overview}
\end{figure}

Seshat is a user-friendly web-based interface whose objective is to smoothly manage large campaigns of audio data annotation, see Figure \ref{fig:overview}. 
Below, we describe the several terms used in Seshat's workflow:

\begin{description}[font=\bfseries, leftmargin=1cm, style=nextline]
    \item[Audio Corpus]
    A set of audio/speech files that a \textbf{Campaign Manager} wants to annotate. It is indicated either by a folder containing sound files, or by a CSV summarizing a set of files. We support the same formats as Praat so far: WAV, Flac and MP3.
    
    \item[Annotation Campaign]
    An object that enables the \textbf{Campaign Manager} to assign \textbf{Annotation Tasks} to the \textbf{Annotators}. It references a \textbf{Corpus}, and allows the Manager to track the annotation's tasks progress and completion in real time. At its creation, a \textbf{Textgrid Checking Scheme} can also be defined for that campaign.
    
    \item[Annotation Task]
    It is contained in an \textbf{Annotation Campaign}, it references an audio file from the campaign's designated \textbf{Audio Corpus}, and assigned to \textbf{Annotators}. It can either be a \textit{Single Annotator Task} (assigned to one Annotator) or a \textit{}{Double Annotator Task} (assigned to two annotators, who will annotatote the assigned task in parallel).
    
    \item[Textgrid Checking Scheme]
    A set of rules defining the TextGrid files' structure and content of the annotations. It is set at the beginning of the \textbf{Annotation Campaign's} creation, and is used to enforce that all TextGrids from the campaign contain the same amount of Tiers, with the same names. It can also enforce, for certain chosen tiers, a set of valid annotations.
    
    \item[Campaign Manager]
    Users with the rights to create \textbf{Annotation Campaigns} and \textbf{Annotators} user accounts, and assign \textbf{Annotation Tasks} to \textbf{Annotators}.
    
    \item[Annotator]
    Users who are assigned a set of \textbf{Annotation Tasks}. Their job is to complete the annotation of the audio files with the Praat software. \par 
    If the TextGrid file they submit does not comply with their \textbf{Annotation Task}'s \textbf{TextGrid Checking Scheme}, Seshat pinpoint their annotation errors with detailed messages. The annotator can re-submit the concerned file to the platform based on these different feedbacks.
\end{description}

Once they they connected to their instance of Seshat, \textit{campaign managers} can access ongoing annotation campaigns or create new ones.  
Campaign managers are able to add \textit{annotators}, assign \textit{annotation tasks} and track progress. Annotator see a list of assigned tasks. The first step for them is to download the sound file with its corresponding auto-generated template TextGrid. In the current implementation, the annotation work has to be done locally with Praat. An upcoming version will use of web tools like Praat on the web \cite{dominguez2016praat}. Once the task is completed, the TextGrid file is to be uploaded to Seshat via the web interface. We used the TextGrid format because of the wide acceptance of the Praat software in the speech science community (e.g., language acquisition research, clinical linguistics, phonetics and phonology). \par

The \textit{Textgrid Checking Scheme} that encompasses rules on the tier's naming, file structure, and the content of the annotations, is associated with a specific campaign and defined at the creation of the campaign. Seshat back-end will automatically check that the submitted TextGrid file  conforms to the \textit{Annotation Campaign}'s \textit{Textgrid Checking Scheme}. \par

\begin{figure*}[!ht]
  \center
  \includegraphics[width=\textwidth]{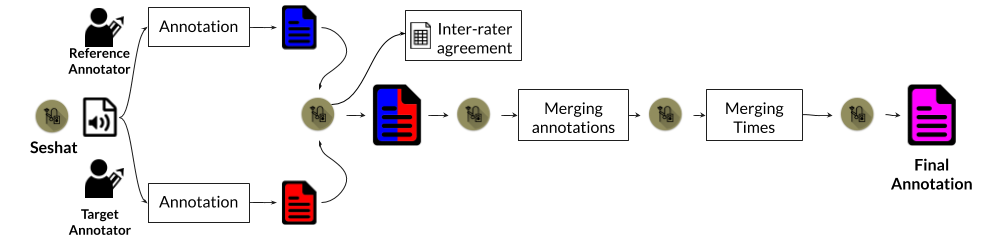}
  \caption{Double Annotator task overview. Inter-rater agreement is computed by the interface for the first independently annotated files in Red and Blue.}
  \label{fig:double}
\end{figure*}

Seshat allows the campaign manager to create two type of tasks: single annotator, and double annotator.   
Regarding the first task, one audio file is attributed to one annotator. Once the annotation is completed, Sesha automatically checks the conformity of the annotation, and only declares a tasks completed if the conformity checks is passed.   Regarding the second task, one audio file is attributed to two annotators. The two annotators annotate the same file \emph{independently}, then the two versions are merged and the annotators are guided through a \emph{compare and review} process to agree one final version. We summarise in the Figure \ref{fig:double} the different steps for the double-annotator task. At each step during merging, the two annotators are provided feedbacks to focus on where are the disagreements.
This process also results in the computation of an \emph{Inter-annotator agreement} for each file. The double annotator task can be used to train new annotators alongside experts. 

Annotating speech data is a joint task of segmentation and categorisation of audio events. That is why we adopted the $\gamma$ measure \cite{mathet2015unified} to evaluate the inter- or intra- annotator agreement in each individual tier. Campaign manager can customise the distance used by $\gamma$ by inserting a custom distance along their own parser (See short snippet of code for a parser of French Phonetics with the SAMPA alphabet in Algorithm \ref{algo:sampa_parser}).

\section{Development}
\label{main:dev}
\subsection{Engineering choices}

Our utmost priority when building Seshat was to make it as easy as possible for others to deploy, use, administer and eventually contribute to.
To do so, we chose the most common frameworks that are free and open-source, all of which are detailed in the following sections. 
Additionally, to match the current trend in web development, we decided to use the so-called "web-app" architecture for Seshat, i.e., we separated the application into two distinct entities: a \emph{front-end}, running on the browser, and a \emph{back-end}, serving data to the front-end and interacting with the database.

\subsubsection{Back-end Choices}
The back-end system runs on a server. It holds and updates the campaign databases and runs the annotation checking and inter-rater agreement evaluation services.  We chose Python, given its widespread use in the scientific community\footnote{\url{https://insights.stackoverflow.com/survey/2019}}, with a wide array of speech and linguistic packages. Moreover, its usage on the back-end side will allow the future integration of powerful speech processing tools like Pyannote \cite{bredin2019pyannote} to semi-automatize  annotations. We thus went for Python3.6 for Seshat's server back-end. We used the Flask-Smorest\footnote{\url{https://github.com/marshmallow-code/flask-smorest}} extension (which is based on Flask\footnote{\url{https://www.palletsprojects.com/p/flask/}}) to clearly and thoroughly document our API, which can be exported to the popular OpenAPI 3.0.2\footnote{\url{https://github.com/OAI/OpenAPI-Specification/blob/master/versions/3.0.2.md}} RESTful API description format.

The files and server data are stored on a MongoDB\footnote{\url{https://www.mongodb.com/}} database, chosen for its flexible document model and general ease of use. We used the Object-Relational Mapping (ORM) MongoEngine\footnote{\url{http://mongoengine.org/}} to define our database schemas and interact with that database. 
MongoDB's GridFS system also allowed us to directly store annotation files (which are usually very light-weight) directly in the database, instead of going through the file system.

\subsubsection{Front-end Choices}
The front-end handles all of the interactions between the users (campaing manager or annotator) with the databses. It is implemented as an App within their browser. We decided to base Seshat's front-end on the Angular Typescript \footnote{\url{https://angular.io/}} framework. Despite its' steep learning curve, it enforces strict design patterns that guarantee that others can make additions to our code without jeopardising the stability of the App. Angular Typescript has a wide community support in the web development industry and is backed by Google and Microsoft. Moreover, the fact that it is based on TypeScript alleviates the numerous shortcomings of JavaScript, ensuring our implementation's readability and stability.

\subsection{UX/UI Choices}
\begin{figure}[!ht]
  \includegraphics[width=0.48\textwidth]{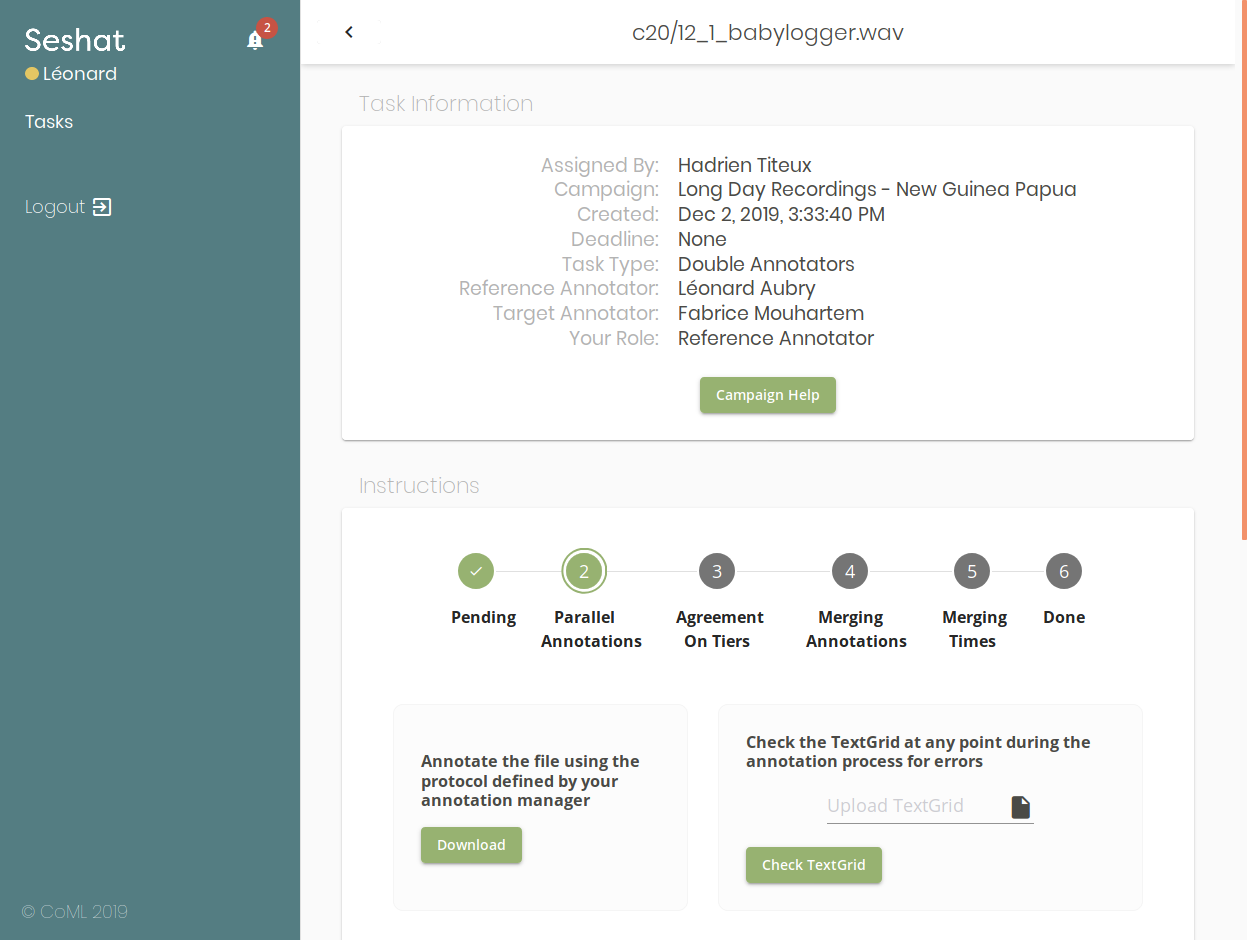}
  \caption{Assigned task from the annotator's point of view.}
  \label{fig:progress_annotator}
\end{figure}

The interface and the features we selected for our implementation are the process of a year-long iterative process involving a team of annotators, two campaign managers and software engineers. We followed some guiding principles from the recent Material\footnote{\url{https://material.io/design/introduction/\#principles}} design language. 
Our goal while designing our interface (with the help of a professional designer) was to make it fully usable by non-technical people.
We also put some extra care into the annotators' interface to give them a clear sense of what is to be done, how they should follow the annotation protocol, and how to correct potential errors in their annotations (See Figure \ref{fig:progress_annotator}) The goal was to reduce the number of actions to perform for annotators and enable to focus only on the annotations content.

\section{Using Seshat}
\label{main:using}
\subsection{Installation and Setup}
Setting up a modern fully-fledged web service is a arduous task, usually requiring a seasoned system administrator as well as sometimes having very precise system requirements. Luckily, the Docker\footnote{\url{https://www.docker.com/}} virtualisation platform ensures that anyone with a recent-enough install of that software can set up Seshat in about one command (while still allowing some flexibility via a configuration file). For those willing to have a more tightly-controlled installation of Seshat on their system, we also fully specify the manual installation steps in our \href{https://seshat-annotation.readthedocs.io/}{online documentation}\footnote{\url{https://seshat-annotation.readthedocs.io/}}).

Importing an audio corpus that you are willing to annotate is easy as dropping files into a default `corpora/` folder. It is possible to either drop a folder containing audio files (with no constraints on the folder's structure), or a CSV file listing audio filenames along with their durations (in case the files are sensitive and you're not willing to risk them being hosted on the server). It is then possible to review the automatically imported files \textit{via} the web interface.

\subsection{Launching and monitoring an annotation campaign}

The Campaign manager can easily define and monitor annotation campaign. As shown in Figure~\ref{fig:campaing_creation}, the online form enable to choose corpora, pre-define and pre-configure the annotations scheme (tiers and parsers). There are 2 types of tiers already implemented by default: one with no check at all, and one with pre-defined categories. For the latter, these categories are pre-defined when the campaign is created.

\begin{figure}[!ht]
  \includegraphics[width=0.48\textwidth]{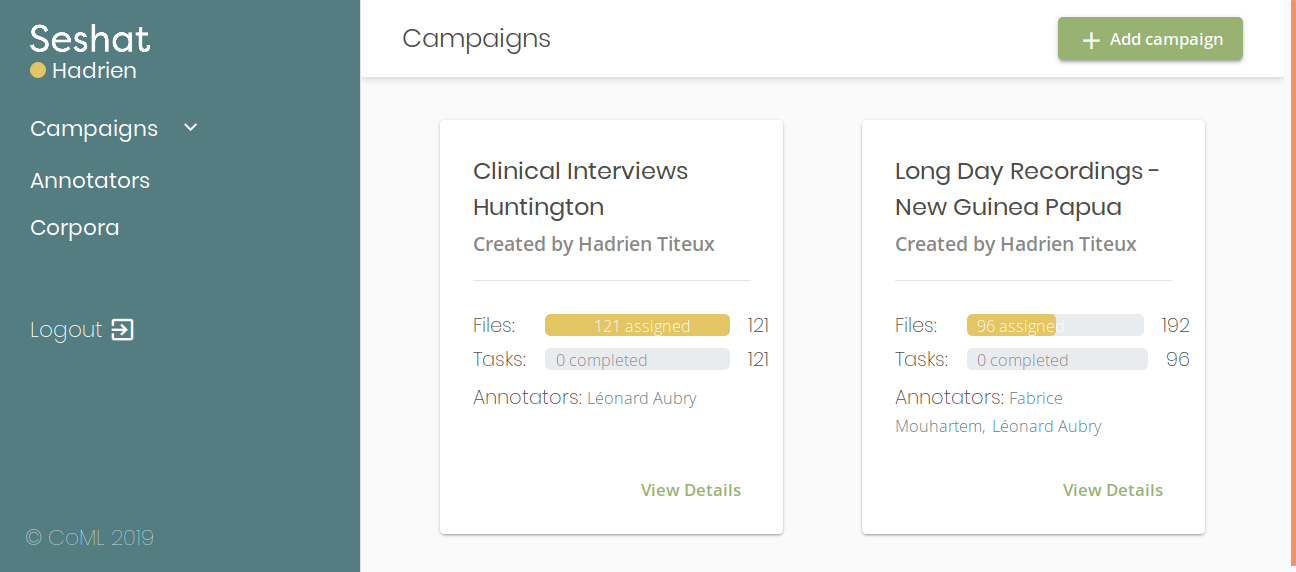}
  \caption{Dashboard for the campaign manager}
  \label{fig:campaigns_overview}
\end{figure}

Only Campaign managers can access and build new campaigns. If Campaign manager have several campaigns they can easily switch between them via the menu bar or get a full overview with the dashboard (See Figure \ref{fig:campaigns_overview}). The campaign managers can visualise the progress of the assigned tasks at the campaign level or more precisely at the task level. They can retrieve all the intermediate files that have been created for each task. For instance, the campaign manager can examine qualitatively and quantitatively what are the annotation differences before the merge phases of the double annotator task.

\subsection{Scripting API}

For those willing to interact with Seshat using code, it is possible to interact with Seshat using either its RESTful API or its command-line interface (CLI). The API endpoints that can be called are all listed in a simple interface, and can be made from any programming language able to make HTTP requests. The CLI interface can be used via your terminal, and therefore can be interacted with using Bash scripts.\par

A typical usage of these features would be to assign annotation tasks from a large speech corpus (spoken by several speakers) to a large pool of annotators, all the while making sure each annotator has a similar number of tasks, with each speaker being evenly distributed among annotators as well. This would be tedious to do manually via the user interface, but easy to program in any scripting language.

\subsection{Annotation Parser Customisation}

We aimed at a reasonable trade-off between simplicity and flexibility for the TextGrid annotations checking component. However, we understand (from our own experience in particular) that sometimes annotations can follow a very specific and complex standard (for instance, parsing SAMPA phonemes strings). To allow users to define their own annotation standards, we added the possibility for users to define an annotation parser, via a simple package-based extension system (taking inspiration from pyannote's extension system). Anyone willing to create a new annotation parser has to be able to program in Python and have a minimal understanding of its packaging system.\par 

As presented in our example French SAMPA Parser (Algorithm~\ref{algo:sampa_parser}), implementing a custom annotation parsers only requires the overload of two methods from Seshat's \texttt{BaseCustomParser} class: 
\begin{itemize}
    \item \texttt{check-annotation}: takes an annotation string as input and raises an error if and only if the annotation is deemed to be invalid. It doesn't return anything.
    \item \texttt{distance}: takes two annotations as input and should return a float corresponding to the distance between these two annotations.
\end{itemize}

\begin{lstlisting}[float=ht, language=Python, caption=Parser Plugin Example. This parser checks the units to allow only phone sequences in the SAMPA format. The distance that can be used for the inter-rater agreement is the Levenshtein distance., label={algo:sampa_parser}, basicstyle=\tiny]
class FrenchSAMPAParser(BaseCustomParser):
    PHONEMES = ['&/', '2','9','9~','@','A','A/','E','E/','H',
                'N','O','O/','R','S','U~/','Z','a','a~','b',
                'd','e','e~','f','g','i','j','k','l','m','n',
                'o','o~','p','s','t','u','v','w','y','z','J']

    def __init__(self):
        # sorting (descending) phonemes by length
        self.PHONEMES = sorted(self.PHONEMES, key=len, reverse=True)

    def parse_sampa(self, pho_str: str) -> List[str]:
        """Parses a French phoneme string into a phoneme list
         ex : "septa~br" -> [s, e, p, t, a~, b, r]"""
        original_str = str(pho_str)
        pho_list = []
        while pho_str:
            parsed_phoneme = False
            # trying to match longest phoneme names first
            for phoneme in self.PHONEMES:
                if pho_str[:len(phoneme)] == phoneme:
                    pho_list.append(pho_str[:len(phoneme)])
                    pho_str = pho_str[len(phoneme):]
                    parsed_phoneme = True
                    break
            if not parsed_phoneme:
                raise AnnotationError(
                    "Can't parse phonetic form %s (stuck at %s)"
                                      % (original_str, pho_str))
        return pho_list

    def check_annotation(self, annot: str) -> None:
        self.parse_sampa(annot)

    def distance(self, annot_a: str, annot_b : str) -> float:
        """Computes the levenshtein distance between two phone sequences"""
        parsed_a = self.parse_sampa(annot_a)
        parsed_b = self.parse_sampa(annot_b)
        return levenshtein(parsed_a, parsed_b)
\end{lstlisting}

\subsection{Inter-rater agreement: the $\gamma$ measure}
It is necessary have a measure of confidence to obtain high-quality datasets and therefore to draw valid conclusions from annotations. Annotations tasks of audio and speech data usually have some specificities. The items to annotate have to be both segmented in time and categorised. The segments can be hierarchically defined or overlapping. In addition, the audio stream may require only sparse annotations (especially in-the-wild recordings which contain a lot of non-speech segments). To evaluate speech annotations, the measure needs to take these characteristics into account. That is why we decided to re-implement and compute the $\gamma$ measure (see \newcite{mathet2015unified} for its design and the advantages of this measure over previous agreement measures).

First, the $\gamma$ software aligns (tier-wise) the annotations of the different annotators. To align the two sets of annotations the $\gamma$ measure the distance between all the individual units. 
The difference of position of two annotated units $u$ and $v$ is measured with the positional distance:
\begin{small}
\[
d_{\text{pos}}(u, v)=\left(\frac{|s t a r t(u)-\operatorname{start}(v)|+|\operatorname{end}(u)-\operatorname{end}(v)|}{(e n d(u)-\operatorname{start}(u))+(e n d(v)-\operatorname{start}(v))}\right)^{2} 
\]
\end{small}

If the tiers are categorical, the distance for the content of the annotated units $u$ and $v$ is defined as:
\begin{small}
\[
d_{\text{cat}}(u, v)= \mathds{1}(cat(u) == cat(v))
\]
\end{small}

This distance can be over-written by the custom parser as mentioned above.
These two distance are summed with equal weights to obtain the distance between every annotated units from 2 annotators. Then, it is possible to obtain the \textit{disorder} $\delta(a)$ of a specific alignment $a$ by summing the distance of all the aligned units in $a$. All possible alignments $a$ are considered and the one that minimises the disorder $\delta(a)$ is kept. 

To get the value of $\gamma$, the disorder is chance-corrected to obtain an expected disorder. It is obtained by re-sampling randomly the annotations of the annotators. 
This means that real annotations are drawn from the annotators, and one position in the audio is randomly chosen. The annotation is split at this random position and the two parts are permuted. It is then possible to obtain an approximation of the expected disorder $\delta_e$. The final agreement measure is defined as:
 
 \[
 \gamma=1-\frac{\delta(a)}{\delta_{e}}
 \]


This $\gamma$ measure is automatically computed by the back-end server for the double-annotator tasks. The Campaign manager can retrieve these measures in Seshat by downloading a simple CSV file.

\section{Use cases}
\label{main:use_cases}
We present two use cases on which Seshat was developped: clinical interviews, and daylong child-centered recordings. 

\subsection{Clinical interviews}
Seshat was intially developped to study the impact of Huntington's Disease \cite{walker2007huntington} on speech and language production. One hundred and fifty two interviews between a neuropsychologist and a patient with the Huntington's Disease (HD) were recorded between June 2018 and November 2019. The campaign manager created a campaign with multiple tiers to annotate the turn takings and the speech/non speech boundaries of the utterances of the patient. For both tasks, the annotations did not need to cover completely the audio (sparsity property mentioned above). For the Turn-taking annotations, there are 3 pre-defined tiers, each one with a single class ('Patient', 'Non-Patient', and 'Noise'), which results in possible overlap between these classes. For the Utterance annotations, there is only one pre-defined class ('Utterance').

\begin{figure}[!ht]
  \includegraphics[width=0.48\textwidth]{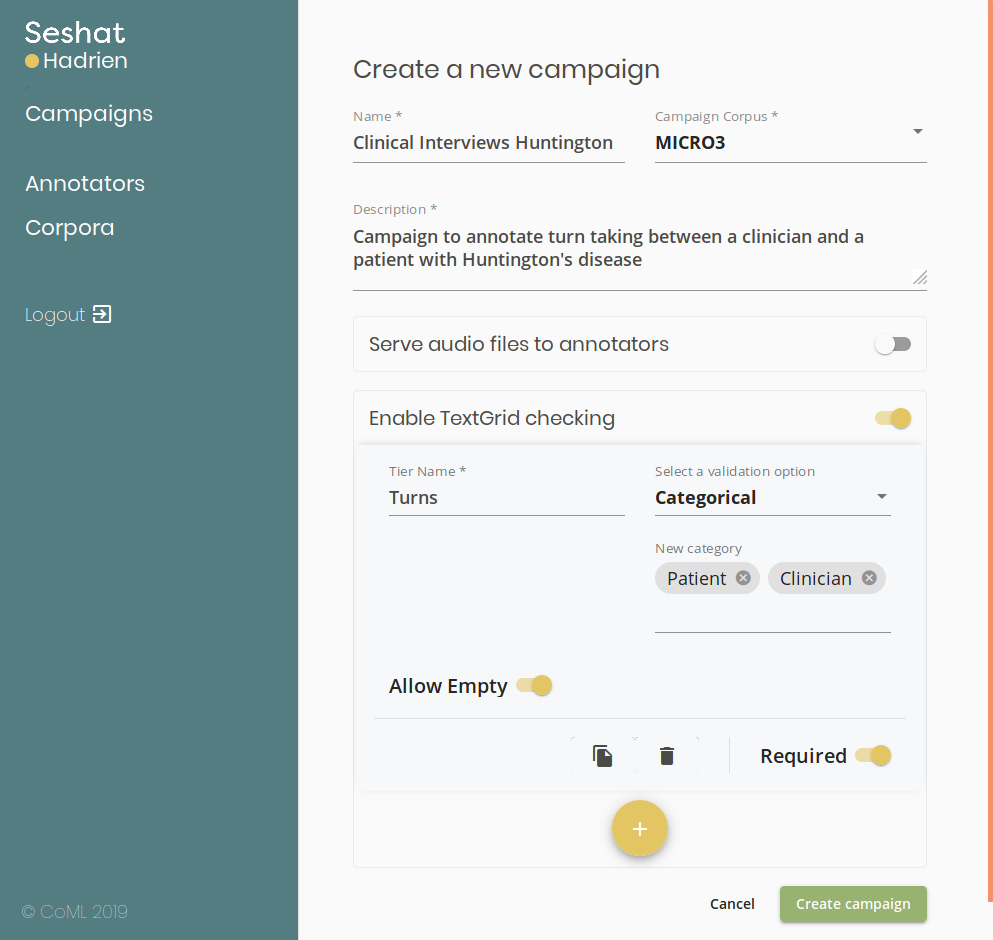}
  \caption{Annotation Campaign definition in Seshat for clinical interviews between a patient with the Huntington's Disease and a neuropsychologist}
  \label{fig:campaing_creation}
\end{figure}

To this date, a total of $67$ files have been fully annotated with the help of Seshat by a cohort of $18$ speech pathologist students (see Figure \ref{fig:campaing_creation}). Among these, $16$ have been done by 2 different annotators independently with the Double-annotator task.  The results are summarised in Table~\ref{tab:hd_gamma}.

Even though there are more categories for Turn-Takings than Utterance (\newcite{gut2004measuring} reported that the more categories the more the task is difficult in speech annotations), the mean $\gamma$ for the Turn-Takings $\gamma= 0.64$ is slightly higher than the one for Utterance $\gamma = 0.61$. And the range of values for the Turn-Takings is smaller than the Utterance. 
Indeed, the speech pathologists reported the difficulty to annotate the boundary of utterances in spontaneous speech, with several ambiguous cases due to pauses. These results will help us to redefine the protocol and be more precise on the given instructions.

\begin{table}[ht]

	\centering
	\begin{tabular}{p{0.25\linewidth}|p{0.10\linewidth}p{0.1\linewidth}p{0.10\linewidth}p{0.15\linewidth}} 
		\hline
		
		Tiers & \multicolumn{3}{c}{$\gamma$} \\
		\hline
				\hline
		               & Mean & Range  & \#classes          \\
		\hline
		
		Turn-Takings               & 0.64 & 0.18  & 3       \\
		
		Utterance              & 0.61    & 0.39 & 1        \\

		\hline

	\end{tabular}
	\caption{$\gamma$ Inter-rater agreements summary for 16 clinical interviews between a neuropsychologist and a patient with the HD. }
	    \label{tab:hd_gamma}

\end{table}

\subsection{In-the-wild child-centered recordings}

The Seshat software is also currently used to annotate audio files in a study of day-long audio-recordings captured by two devices  (LENA \cite{gilkerson2008lena}, and a BabyCloud baby-logger device) worn by young children growing up in remote Papua New Guinea. The project aims at establishing language input and outcomes in this seldom-studied population. To establish reliability levels, 20 1-min files were double-annotated by 2 speech pathology students. Among the tasks given to the annotators there was: (1) locating the portions of Speech (Speech activity), (2) locating the speech produced by an adult that is directed to a child or not (\textit{Adult-Directed Speech} versus \textit{Child-Directed Speech}). As in the previous example, the annotations do not need to cover the full audio file. The Speech Activity task has only 1 class ('Speech') and the Addressee task has 2 classes ('ADS', 'CDS').

\begin{table}[ht]

	\centering
	\begin{tabular}{p{0.28\linewidth}|p{0.10\linewidth}p{0.1\linewidth}p{0.10\linewidth}p{0.15\linewidth}} 
		\hline
		
		Tiers & \multicolumn{3}{c}{$\gamma$} \\
		\hline
				\hline
		               & Mean & Range  & \#classes          \\
		\hline
		
		Speech activity              & 0.46 & 0.60  & 1      \\
		
		ADS vs CDS              & 0.27    & 0.39 & 2        \\

		\hline

	\end{tabular}
	\caption{$\gamma$ Inter-rater agreement for 20 1-min slices extracted from child-centered day-long recordings. ADS and CDS stand for Adult-Directed Speech and Child-Directed Speech respectively.}
	    \label{tab:darcle_gamma}

\end{table}

These recordings have been done in naturalistic and noisy conditions; moreover, the annotators do not understand the language. Probably as a result of these challenges,  agreement between annotators is lower than in the Clinical interviews use case. This information is nonetheless valuable to the researchers, as it can help them appropriately lower their confidence in the ensuing speech quantity estimates.

\section{Conclusion and Future work}
\label{main:conclusion}
Seshat is a new tool for the management of audio annotation efforts. Seshat enables users to define their own campaign of annotations. Based on this configuration, Seshat automatically enforces the format of the annotations returned by the annotators. Besides, we also add the capability to finely tailor the parsing of the annotations. 
Finally, Seshat provides automatic routines to compute the inter-rate agreements that are specifically designed for audio annotations. 
Seshat lays some foundations for more advanced features, either for the interface or the annotation capabilities. In future work, we plan  to implement an automatic task assignments and an integration of a
diarization processing step to reduce human effort. Another planned feature is to add possibility for the campaign manager to design more complex annotation workflows such as, for instance, dependencies between tiers or more intermediate steps of annotations. 

\section{Acknowledgements}
This research was conducted thanks to 
Agence Nationale de la Recherche (ANR-17-CE28-0007 LangAge, ANR-16-DATA-0004 ACLEW, ANR-14-CE30-0003 MechELex, ANR-17-EURE-0017, ANR-10-IDEX-0001-02 PSL*, ANR-19-P3IA-0001, ) and grants from Facebook AI Research (Research Grant), Google (Faculty Research Award), and Microsoft Research (Azure Credits and Grant), and a J. S. McDonnell Foundation Understanding Human Cognition Scholar Award. 
\newpage

\section{References}
\label{lr:ref}

\bibliographystyle{lrec}
\bibliography{bibliography}


\end{document}